\documentclass[conference]{IEEEtran}
\IEEEoverridecommandlockouts
\usepackage{url}
\usepackage{cite}
\usepackage{amsmath,amssymb,amsfonts}
\newtheorem{thm}{Theorem}
\newtheorem{lem}{Lemma}
\newtheorem{prop}{Proposition}
\newtheorem{defn}{Definition}
\newtheorem{exmp}{Example}
\newtheorem{rem}{Remark}
\usepackage{algorithmic}
\usepackage{graphicx}
\usepackage{textcomp}
\def\BibTeX{{\rm B\kern-.05em{\sc i\kern-.025em b}\kern-.08em
		T\kern-.1667em\lower.7ex\hbox{E}\kern-.125emX}}
\usepackage{hyperref}
	
\usepackage{fancyhdr}	

\begin{document}
	
	\title{Robust Maneuver Planning With Scalable Prediction Horizons: A Move Blocking Approach}
	\author{Philipp Schitz, Johann C. Dauer and Paolo Mercorelli
		\thanks{This work was supported by the German Federal Ministry for Economic Affairs and Climate Action under Grant 03EE3059B. Recommended by Senior Editor L. Zhang. (Corresponding author: Philipp Schitz.) }
		\thanks{P. Schitz and J. C. Dauer are with the Institute of Flight Systems, German Aerospace Center (DLR), 38108 Braunschweig, Germany. P. Schitz is also with the Institute for Production Technology and Systems, Leuphana University of Lueneburg, 21335 Lueneburg, Germany (e-mail: philipp.schitz@dlr.de; johann.dauer@dlr.de).}
		\thanks{P. Mercorelli is with the Institute for Production Technology and Systems, Leuphana University of Lueneburg, 21335 Lueneburg, Germany (e-mail: paolo.mercorelli@leuphana.de).}}
	
	\maketitle

	\pagestyle{fancy}
	
	\begin{abstract}
		Implementation of Model Predictive Control (MPC) on hardware with limited computational resources remains a challenge. Especially for long-distance maneuvers that require small sampling times, the necessary horizon lengths prevent its application on onboard computers. In this paper, we propose a computationally efficient tube-based shrinking horizon MPC that is scalable to long prediction horizons. Using move blocking, we ensure that a given number of decision inputs is efficiently used throughout the maneuver. Next, a method to substantially reduce the number of constraints is introduced. The approach is demonstrated with a helicopter landing on an inclined platform using a prediction horizon of 300 steps. The constraint reduction decreases the computation time by an order of magnitude with a slight increase in trajectory cost.
	\end{abstract}
	
	
	\section{Introduction}
	\label{sec:introduction}
	\IEEEPARstart{M}{odel} Predictive Control (MPC) is a popular control method for handling systems subject to state and input constraints \cite{mayneModelPredictiveControl2014,erenModelPredictiveControl2017}. In regular MPC, a trajectory optimization problem is solved over a fixed receding horizon at every time step. In order to ensure closed-loop stability, the terminal state is usually constrained to lie within an invariant terminal set \cite{mayneConstrainedModelPredictive2000}. However, invariance of the terminal set may be restrictive in certain applications \cite{richardsRobustVariableHorizon2006}. 
	For example, reaching a certain position with a non-zero target velocity can not be encoded as an invariant set. It is well known that finite-time arrival to arbitrary terminal sets can be achieved by varying the prediction horizon length \cite{richardsModelPredictiveControl2003,richardsRobustVariableHorizon2006,shekharRobustVariableHorizon2012,shekharRobustModelPredictive2015,koelnTwoLevelHierarchicalMissionBased2018,farahaniShrinkingHorizonModel2017}.
	Usually, the resulting optimization is a mixed-integer program (MIP) where the horizon length acts as the integer decision variable. Finite-time completion and recursive feasibility require that the target is reachable from the current state. As a result, large horizons are often needed to achieve practically relevant regions of attraction (ROA). In \cite{shekharRobustModelPredictive2015}, the required horizon length is reduced by introducing intermediate waysets. By placing these sets in a way that ensures the reachability from one wayset to another, an unmanned aircraft is shown to maneuver to the target region while avoiding obstacles. However, the resulting MIPs need to be solved online which is usually prohibitively expensive on onboard computers. Another strategy \cite{koelnTwoLevelHierarchicalMissionBased2018,farahaniShrinkingHorizonModel2017}, often referred to as shrinking horizon MPC (SHMPC), treats the horizon length as a parameter so the problem reduces to a continuous program. Consequently, optimality with respect to horizon length is sacrificed for computational tractability. In the aforementioned approaches, the number of decision variables and thus the computational load vary drastically with horizon length. Since an onboard computer needs to be able to handle the worst-case scenario, the maximal horizon is limited by hardware capabilities. The approach by \cite{shekharRobustVariableHorizon2012} allows to set a maximal number of decision inputs, regardless of horizon length, using move blocking \cite{cagienardMoveBlockingStrategies2007}, a scheme where inputs are held constant over some number of steps to reduce the amount of decision variables. The trajectory is then optimized over feasible blocking strategies in a MIP.

	\subsection{Main Contribution and Structure of the Paper}
	In this paper, we present a computationally efficient and robust SHMPC that is scalable to long prediction horizons. In contrast to  \cite{shekharRobustVariableHorizon2012}, we simplify the problem to a shrinking horizon formulation in which the horizon length at a given time step is known a priori. On this basis, we present a constructive method to generate time-varying blocking strategies while efficiently utilizing a given maximum number of decision inputs and ensuring recursive feasibility. For long prediction horizons, the number of constraints becomes another significant source of complexity. We therefore reformulate the standard move blocking optimization problem to enable the reduction of constraints. An optimization-based approach is developed that enables the approximation of the constraint set with substantially fewer halfspaces. The efficacy of our approach is demonstrated on a helicopter landing example. The constraint reduction is shown to decrease the computation time by an order of magnitude.
	
	The rest of this paper is structured as follows. Section \ref{sec:problem_statement_and_preliminaries} introduces the control problem and reviews some preliminaries. In Section \ref{sec:SHMPC_with_MB}, the proposed SHMPC with move blocking is presented. The constraint reduction is developed in Section \ref{sec:computational_aspects}. Section \ref{sec:example} demonstrates the approach on a numerical example. Section \ref{sec:conclusion} provides a conclusion and outlook. 
	
	\subsection{Notation}
	Given two sets $\mathcal{S}_1$, $\mathcal{S}_2$, set addition and difference are defined as $\mathcal{S}_1 \oplus \mathcal{S}_2 = \{a + b \, | \, a \in \mathcal{S}_1, \, b \in \mathcal{S}_2 \}$ and $\mathcal{S}_1 \ominus \mathcal{S}_2 = \{a \, | \, a \oplus \mathcal{S}_2 \subseteq \mathcal{S}_1 \}$, respectively. Given matrices or vectors $A$, we interpret $A\mathcal{S}$ as $\{As\,|\,s \in \mathcal{S}\}$. We further denote a block diagonal stacking by $\mathrm{diag}(A_1,\ldots,A_n)$, a vertical concatenation by $(A_1,\ldots,A_n)$ and the Kronecker product by $A_1 \otimes A_2$.  The $n$-dimensional identity matrix is denoted by $I_n$. We interpret $\mathbf{0}_{n \times m}$ and $\mathbf{1}_{n \times m}$ as $n \times m$-dimensional matrix of zeros and ones, respectively. 
	The set of natural numbers ranging from $l$ to $u$ is written as $\mathbb{N}_l^u$.
	\section{Problem Statement and Preliminaries}\label{sec:problem_statement_and_preliminaries}
	Let $x_k \in \mathbb{R}^n$ denote the state at time $k \tau$ with $k \in \mathbb{N}$ and sample time $\tau$. Consider the linear discrete-time system
	\begin{equation}\label{eq:system}
		\begin{aligned}
			x_{k+1} & = A x_k + B u_k + w_k,
		\end{aligned}
	\end{equation}
	subject to state-input and terminal constraints 
	\begin{equation}\label{eq:sys_constraints}
		\forall k \in \mathbb{N}_0^{N_0-1}: \; (u_k, \, x_k) \in \mathcal{F},\; x_{N_0} \in \mathcal{X}_T,
	\end{equation}
	where $u \in \mathbb{R}^m$ is the input, $w_k \in \mathcal{W} \subseteq \mathbb{R}^l$ is an unknown but bounded disturbance and $N_0 \in \mathbb{N}^+$ is the number of time steps within the maneuver. We assume a maximum number of decision inputs $\bar{N}_\text{max}$ (e.g. based on hardware capabilities). We aim to compute finite-time trajectories with a number of decision inputs $\bar{N} \leq \bar{N}_\text{max}$, irrespective of $N_0$. The cost function to be minimized is of the form 
	\begin{equation}\label{eq:cost}
		J = \sum_{i=0}^{N-1} [x_i, \, u_i]^T H [x_i, \, u_i] + x_N^T P x_N,
	\end{equation}
	where $H$, $P$ are positive definite matrices. We assume that (i) the pair $(A,B)$ is stabilizable and (ii) the sets $\mathcal{F}$, $\mathcal{X}_T$, and $\mathcal{W}$ are polytopes containing the origin. In the following, the fundamentals of tube MPC and move blocking are discussed.
	\paragraph{Tube MPC} In order to deal with uncertainties, tube-based approaches \cite{mayneRobustModelPredictive2005,chisciSystemsPersistentDisturbances2001} provide a way to achieve stability guarantees while keeping the resulting optimization computationally tractable. The concept is based on an ancillary controller $K$ that keeps the states of the uncertain system within a tube centered around the states of a nominal system
	\begin{equation}\label{eq:nominal_system}
		z_{k+1} = Az_k + B v_k.
	\end{equation}
	In \cite{mayneRobustModelPredictive2005}, the tube size is constant and given by a Robust Positively Invariant (RPI) set $\mathcal{Z}$.
	\begin{defn}[Robust positively invariant (RPI) set]
		Let $A_K = A-BK$ be such that $x_{k+1} = A_K x_k$ is stable. A set $\mathcal{Z}$ is robust positively invariant if $A_K \mathcal{Z} \oplus \mathcal{W} \subseteq \mathcal{Z}$.
	\end{defn}
	The following proposition establishes that states starting in this tube will in fact remain inside of it for all time under an appropriate control law.
	\begin{prop}[Proposition 1 in \cite{mayneRobustModelPredictive2005}] \label{prop:RPI}
		Let $\mathcal{Z}$ be a RPI set for system \eqref{eq:system}. If $x_0 \in z_0 \oplus \mathcal{Z}$ and we choose $u_k = v_k - K(x_k - z_k)$,	then $x_{k} \in z_{k} \oplus \mathcal{Z}$ for all $w_k \in \mathcal{W}$ and $k \in \mathbb{N}$.
	\end{prop}
	Based on Proposition \ref{prop:RPI}, it is therefore possible to ensure \eqref{eq:sys_constraints} despite the influence of $w$ by performing the trajectory optimization on the nominal system with tightened constraints 
	\begin{equation}\label{eq:tight_constraints}
		\bar{\mathcal{F}} = \mathcal{F} \ominus (\mathcal{Z} \times K\mathcal{Z}), \; \bar{\mathcal{X}}_T = \mathcal{X}_T \ominus \mathcal{Z}.
	\end{equation}
	The nominal system then acts as a reference for the ancillary controller $K$. By adding a constraint to assure that the measured state $x_k$ lies within the tube $z_0 \oplus \mathcal{Z}$ at the start of the trajectory, we obtain the optimization problem $\mathbb{P}_0(x_k)$:
	\begin{equation} \label{eq:0_MPC}
		\begin{aligned}
			\min_{V,z_0} \quad & \sum_{i=0}^{N-1} (z_i, \, v_i)^T H (z_i, \, v_i) + z_N^T P z_N \\
			\textrm{s.t.}  \quad & x_k \in z_0 \oplus \mathcal{Z}, \; z_N \in \bar{\mathcal{X}}_T \\
			& z_{i+1} = A z_i + B v_i, \; (z_i,v_i) \in \bar{\mathcal{F}}, \; i \in \mathbb{N}_0^{N-1},
		\end{aligned}
	\end{equation}
	where $i$ denotes the prediction time step. With the optimal solution at time $k$ given as $V^*(k) = (v_0^*(k),\ldots,v_{N-1}^*(k))$, $z_0^*(k)$, the control law is
	\begin{equation}\label{eq:u_MPC}
		u_k = v_0^*(k) - K(x_k - z_0^*(k)).
	\end{equation}
	
	\paragraph{Move blocking} The computational complexity of the optimization problem is heavily influenced by the number of decision variables. In regular MPC, this corresponds to the number of decision inputs within the horizon. One strategy to reduce complexity is to hold the inputs constant over a certain number of steps using a blocking matrix.
	\begin{defn}[Blocking matrix]\label{defn:blocking_matrix}
		A blocking matrix $M \in \mathbb{R}^{N \times \bar{N}}$ is defined as \begin{equation*}
			M = \mathrm{diag} (\mathbf{1}_{s_1\times1},\ldots,\mathbf{1}_{s_{\bar{N}}\times1}) =: \mathbf{M}(s),
		\end{equation*}
		where the blocking vector $s = (s_1, \ldots, s_{\bar{N}})$ stores the individual blocking lengths $s_i \in \mathbb{N}^+$. 
	\end{defn}
	The matrix $M$ relates the blocked decision input vector $\bar{U} := (\bar{u}_0,\ldots,\bar{u}_{\bar{N}-1})$ to the original one by $U = (M \otimes I_m ) \bar{U}$. The number of inputs within the optimization is therefore reduced from $Nm$ to $\bar{N}m$. For example, a system with $m=1$ and $N=4$ can be reduced to $\bar{N}=2$ decision inputs via
	\begin{equation*}
		\begin{bmatrix}
			u_1 \\
			u_2 \\
			u_3 \\
			u_4 \\
		\end{bmatrix}= M \begin{bmatrix}
			\bar{u}_1 \\
			\bar{u}_2 \\
		\end{bmatrix} \text{ with } M=\begin{bmatrix}
			1 & 0 \\
			1 & 0 \\
			1 & 0 \\
			0 & 1 \\
		\end{bmatrix}, \, s=(3,1).
	\end{equation*}
	
	\section{Efficient Move Blocking for SHMPC}\label{sec:SHMPC_with_MB}
	In this section, we present a time-varying move blocking scheme for SHMPC that efficiently utilizes a given maximum number of decision inputs, regardless of the horizon length. We then prove finite-time completion and recursive feasibility.
	
	\subsection{Optimization Problem Formulation}
	With move blocking, we can ensure that the number of decision inputs will not exceed a given bound $\bar{N}_\text{max}$ by choosing an appropriate blocking matrix at each time step. In order to achieve robustness against disturbances, we combine move blocking with the tube-based MPC formulation \eqref{eq:0_MPC} to arrive at $\mathbb{P}(x_k,M_k)$:
	\begin{equation} \label{eq:MB_SHMPC}
		\begin{aligned}		
			\;\min_{\bar{V},z_0} \; & \sum_{i=0}^{N_k-1} (z_i, v_i)^T H (z_i, v_i) + z_{N_k}^T P z_{N_k},\\
			\textrm{s.t.}  \; &  x_k \in z_0 \oplus \mathcal{Z}, \; z_{N_k} \in \bar{\mathcal{X}}_T, \\
			& V = (v_0,\ldots,v_{N_k-1}) = (M_k \otimes I_m ) \bar{V}, \\
			& z_{i+1} = A z_i + B v_i,  \\
			& (z_i,v_i) \in \bar{\mathcal{F}}, \; i \in \mathbb{N}_0^{N_k-1},
		\end{aligned}
	\end{equation}
	where $\bar{V} = (\bar{v}_0,\ldots,\bar{v}_{\bar{N}})$ is the reduced input vector, $N_k := N_0 - k$ and the control is given by \eqref{eq:u_MPC}. Since the horizon $N_k$ shrinks at every time step, the blocking matrix $M_k \in \mathbb{R}^{N_k \times \bar{N}}$ and the corresponding blocking vectors $s_k = (s_{1,k}, \ldots, s_{\bar{N},k})$ are now time-varying. Optimal values are denoted by a star as their superscript, e.g. $\bar{V}^*(k)$, ${z}_0^*(k)$.
	Note that while the RPI set $\mathcal{Z}$ ensures invariance with respect to the trajectory tracking error $x_k - {z}_0^*(k)$, invariance of the nominal trajectory $z_i$ is not required due to the use of shrinking horizons \cite{richardsRobustVariableHorizon2006}. In the following, we outline a procedure to obtain blocking matrices that ensure recursive feasibility and efficiently utilize the available number of decision inputs.
	
	\subsection{Recursive Feasibility via Truncation}
	The move blocking scheme is built on the fact that, when using shrinking horizons, a feasible solution at the next time step is easily obtained by truncating the previous solution. 
	\begin{prop}\label{prop:feasibleM}
		Consider \eqref{eq:system} controlled by \eqref{eq:u_MPC}. Let $V^*(k) = (M_k \otimes I_m ) \bar{V}^*(k)$ denote the optimal input trajectory of $\mathbb{P}(x_k,M_k)$ for $k \in \mathbb{N}_0^{N_0-1}$. Let $R = [\mathbf{0}_{(N_k-1) \times 1},I_{N_k-1}] \in \mathbb{R}^{(N_k-1)\times N_k}$ denote a truncation matrix. If there exists an input trajectory $W$ such that
		\begin{equation}\label{eq:shiftedSolution}
			(M_{k+1} \otimes I_m ) W =  (R \otimes I_m ) V^*(k) = (v_1^*,\ldots,v_{N-1}^*),
		\end{equation}
		then $\bar{V}(k+1)=W$ and $z_0(k+1)=z_1^*(k)$ are a feasible solution to $\mathbb{P}(x_{k+1},M_{k+1})$.
	\end{prop}
	\begin{IEEEproof}
		We first show that the truncated input trajectory $(R \otimes I_m ) V^*(k)$ is feasible at $k+1$. Based on Proposition \ref{prop:RPI}, $x_{k+1} \in z_1^*(k) \oplus \mathcal{Z}$. Thus, $z_0(k+1)=z_1^*(k)$ satisfies the initial state constraint at $k+1$. Since there is no disturbance acting on the nominal system \eqref{eq:nominal_system}, truncating the previous input trajectory $V^*(k)$ via $(R \otimes I_m ) V^*(k)$ yields a feasible solution. Therefore, if \eqref{eq:shiftedSolution} holds, then $\mathbb{P}(x_{k+1},M_{k+1})$ is feasible for $\bar{V}(k+1)=W$ and $z_0(k+1)=z_1^*(k)$.
	\end{IEEEproof}
	If such a $W$ exists for a particular blocking matrix $M_{k+1}$, we refer to $M_{k+1}$ as a feasible blocking matrix. A simple way to construct a feasible blocking matrix is to choose $W = \bar{V}^*(k)$ and then truncate the first row of $M_k$ as follows:
	\begin{equation} \label{eq:shift}
		\begin{aligned}
			(R \otimes I_m ) V^*(k) &= (R \otimes I_m )(M_k \otimes I_m ) \bar{V}^*(k) \\
			&= (RM_k \otimes I_m)\bar{V}^*(k),
		\end{aligned}				
	\end{equation}
	where we used $(A \otimes B)(C \otimes D) = AC \otimes BD$ for matrices $A,B,C,D$ of appropriate size. Thus, choosing $M_{k+1} = RM_k$ results in a feasible blocking matrix. However, repeatedly truncating $M$ eventually leads to zero columns, effectively nullifying the corresponding reduced input. Consequently, the available number of decision inputs is not fully utilized.
	
	\subsection{Recursively Feasible Interval Splits}
	
\begin{figure}
		\centering
		\includegraphics[width=0.49\textwidth]{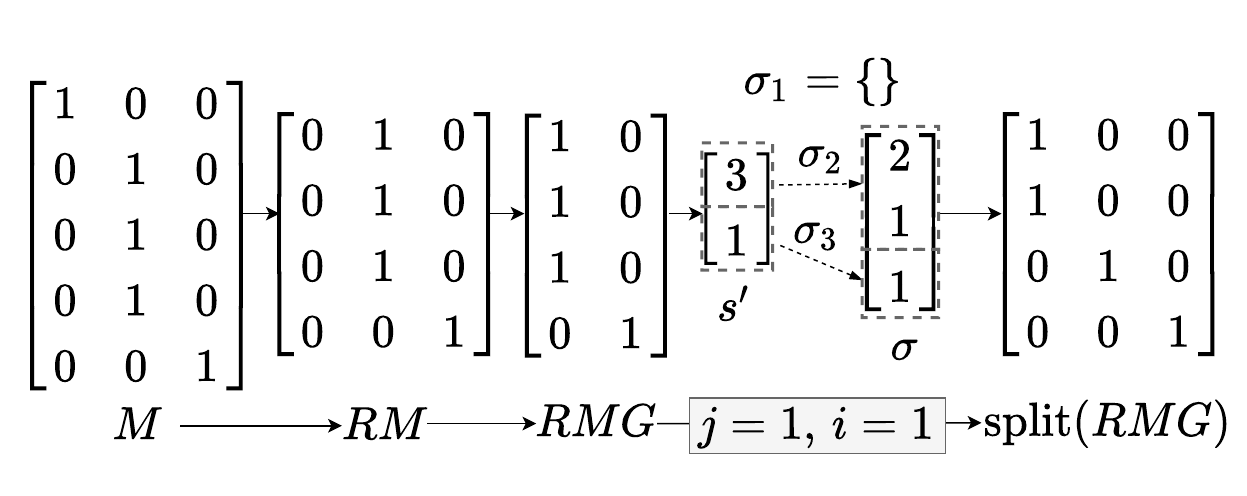}
		\caption{An example of \eqref{eq:Psi} with $s = (1,3,1)$. The blocking vector $s'$ corresponds to the blocking matrix $RMG$. Since $j=1$, $\sigma_1$ is empty.}
		\label{fig:MoveblockingExample2}
	\end{figure}

	In this section, we expand the previous strategy to enable the efficient usage of the available decision inputs. Our goal is to design a function $\Gamma(M_k)$ that only behaves like $RM_k$ when $s_{1,k} > 1$, i.e. when $RM_k$ does not lead to a reduction of effective decision inputs. However, when $s_{1,k} = 1$ and the horizon length is larger than the maximum number of available decision inputs $\bar{N}_{\mathrm{max}}$, we split an existing blocking interval instead of truncating it. The number of effective decision inputs thus remains $\bar{N}_{\mathrm{max}}$ instead of reducing to $\bar{N}_{\mathrm{max}}-1$. Specifically, let
	\begin{equation}\label{eq:Gamma}
		\Gamma(M_k) = \begin{cases}
			\Psi(M_k)&\text{ if } s_{1,k} = 1 \text{, }N_k > \bar{N}_{\mathrm{max}}  \\
			RM_k &\text{ otherwise}
		\end{cases},
	\end{equation}
	where
	\begin{gather}\label{eq:Psi}
		\Psi(M_k) = \mathsf{split}(RM_kG),\; G = [\mathbf{0}_{(\bar{N}-1)\times1}, I_{\bar{N}-1}]^T.
	\end{gather}
	An example of \eqref{eq:Psi} is shown in Fig. \ref{fig:MoveblockingExample2}. The function $\Psi(M_k)$ first truncates $M_k$ using $R$. Since $s_{1,k} = 1$, removing the first row results in a matrix with only zeros in the first column. This column is removed using $G$, leading to a blocking matrix with $\bar{N}-1$ columns. The result is then split using the function $\mathrm{split}$, which we define in the following. Recalling Definition \ref{defn:blocking_matrix}, $M$ can be represented using its blocking vector $s$ via $M = \mathbf{M}(s)$. Our proposed splitting procedure can be written as:
	\begin{align}\label{eq:split}	
		\mathrm{split}(M) = \mathbf{M}(\sigma),\quad \sigma = (\sigma_1,\sigma_2,\sigma_3)
	\end{align}
	where $\sigma_1 = (s_1,\ldots,s_{j-1})$, $\sigma_2 = (s_j-i,i)$, and $\sigma_3 = (s_{j+1},\ldots,s_{\bar{N}})$ with $i \in \mathbb{N}_1^{s_j-1}$ and $j \in \mathbb{N}_1^{\bar{N}}$ s.t. $s_j >1$. The index $j$ is restricted to columns with $s_j >1$ since a column with $s_j = 1$ can not be split further. Note that for $j = 1$ and $j=\bar{N}$, the vectors $\sigma_1$ and $\sigma_3$ are empty, respectively.
	The following Proposition establishes that $\mathrm{split}(M)$ results in a feasible blocking matrix.
	\begin{prop}\label{prop:feasibleSplits}
		For a feasible blocking matrix $M \in \mathbb{R}^{N \times \bar{N}}$, the blocking matrix $\mathrm{split}(M) \in \mathbb{R}^{N \times \bar{N}+1}$ defined in \eqref{eq:split} is feasible.
	\end{prop}
	\begin{IEEEproof}
		By conjoining columns $j$ and $j+1$ of $\mathrm{split}(M)$, the original blocking matrix $M$ can be recovered. Therefore, $M = \mathrm{split}(M) D_j$
		with $D_j = \mathrm{diag}(I_{j-1},(1,1),I_{\bar{N}-j})$. Since $M$ is feasible, Proposition \ref{prop:feasibleM} states that there exists a feasible input trajectory $(M\otimes I_m) \bar{V}$ so that
		\begin{equation}
			\begin{aligned}
				(M\otimes I_m) \bar{V} &= ( \mathrm{split}(M) D_j \otimes I_m) \bar{V}\\
				&= ( \mathrm{split}(M) \otimes I_m) ( D_j \otimes I_m) \bar{V}.
			\end{aligned}		
		\end{equation}
		It follows that $( D_j \otimes I_m) \bar{V}$ can be chosen as a feasible input trajectory for $\mathrm{split}(M)$, thus it is a feasible blocking matrix.
	\end{IEEEproof}
	Proving recursive feasibility and finite time completion is now straightforward.
	\begin{thm}\label{thm:recursive_feasibility}
		Consider \eqref{eq:system} controlled by \eqref{eq:u_MPC}, wherein $v_0^*(k)$ and $z_0^*(k)$ are part of the optimal solution to $\mathbb{P}(x_k,\Gamma(M_{k-1}))$. Suppose that a solution to $\mathbb{P}(x_0,M_0)$ exists for some $M_0$ with an initial horizon length $N_0$. Then
		\begin{enumerate}
			\item $\mathbb{P}(x_k,\Gamma(M_{k-1}))$ is feasible for all $k \in \mathbb{N}_1^{N_0-1}$;
			\item $x_{N_0} \in \mathcal{X}_T$.
		\end{enumerate}	
	\end{thm} 
	\begin{IEEEproof}
		Suppose that $M_{k-1}$ is feasible for some $k \in \mathbb{N}_1^{N_0-1}$. If $M_k = \Gamma(M_{k-1})$ is feasible, item 1) holds by induction since the base case is true by assumption. For the case $N_{k-1} > \bar{N}$ and $s_{1,k-1} = 1$ of \eqref{eq:Gamma}, we first need to show that $M'_k := RM_kG$ in \eqref{eq:split} is feasible in order for Proposition \ref{prop:feasibleSplits} to be applicable. Since the first column of $RM_{k-1}$ consists of zeros, we can add a zero column to $M'_{k-1}$ to obtain $RM_{k-1} = [\mathbf{0}_{(N_{k-1}-1)\times1}, M'_{k-1}] = M'_{k-1} G^T$. As \eqref{eq:shift} shows that is $RM_{k-1}$ feasible, we obtain 
		\begin{multline}
			(RM_{k-1} \otimes I_m)\bar{V}^*(k-1) \\
			\begin{aligned}
				&= (M'_{k-1} G^T \otimes I_m) \bar{V}^*(k-1) \\
				&= (M'_{k-1} \otimes I_m) ( G^T \otimes I_m) \bar{V}^*(k-1),
			\end{aligned} 
		\end{multline}
		i.e. $M'_k$ is feasible with $W = (G^T \otimes I_m) \bar{V}^*(k-1)$ according to Proposition \ref{prop:feasibleM}. By Proposition \ref{prop:feasibleSplits}, $\mathrm{split}(M'_{k-1})$ is then also feasible. For all other cases, we have $M_k = RM_{k-1}$ which is again shown to be feasible in \eqref{eq:shift}. $\Gamma(M_{k-1})$ is thus feasible any $k \in \mathbb{N}_1^{N_0-1}$. Item 2) follows directly: At time step $k = N_0$, since $\mathbb{P}(x_k,\Gamma(M_{k-1}))$ is feasible for all $k \in \mathbb{N}_1^{N_0-1}$, Proposition \ref{prop:RPI} guarantees that $x_{N_0} \in z_1^*(N_0-1) \oplus \mathcal{Z} \in \mathcal{X}_T$.
	\end{IEEEproof}
	\section{Constraint Reduction}\label{sec:computational_aspects}	
	For move blocking, the feasibility of the state trajectory between two blocked inputs is ensured by encoding that every state within the interval satisfies the constraints. A key observation is that this kind of formulation leads to highly redundant constraints and thus increases the complexity of the problem unnecessarily. In this section, we reformulate $\mathbb{P}$ to allow for constraint reduction and then present an approach to efficiently approximate the constraint sets. This facilitates the scalability of our SHMPC with respect to horizon length.
	
	\subsection{Transition Matrix Formulation}
	Let $\xi_0(i) = (z_i, \bar{v}_i) \in \bar{\mathcal{F}}$ denote the state-input pair at the start of a blocking interval $s_{i,k}$ ($k$ is dropped in the following for brevity). The following auxiliary system describes the trajectories evolving within this interval:
	\begin{align}\label{eq:aug_sys}
		\xi_{j+1}(i) = \mathbf{A} \xi_j(i), \; \mathbf{A} = \begin{bmatrix} A & B \\ \mathbf{0}_{m \times n} & I_m \end{bmatrix}.
	\end{align}
	With $\tilde{A}(s_i) = A^{s_i}$ and $\tilde{B}(s_i) = \sum_{j=0}^{s_i-1} A^j B$, we then have $
	z_{i+s_i} = \tilde{A}(s_i) z_i + \tilde{B}(s_i) v_i$. Therefore, \eqref{eq:MB_SHMPC} can be equivalently stated as $\tilde{\mathbb{P}}(x_k,s_k)$:
	\begin{equation}
		\begin{aligned}
			\min_{\tilde{z}_0,\bar{V}} \quad & \sum_{i=0}^{\bar{N}-1} (\tilde{z}_i, \, \bar{v}_i)^T \tilde{H}(s_i) (\tilde{z}_i, \, \bar{v}_i) + \tilde{z}_{\bar{N}}^T P \tilde{z}_{\bar{N}} \\
			\textrm{s.t.} \quad & \tilde{z}_0 \in x_k \oplus \mathcal{Z},\; \tilde{z}_{\bar{N}} \in \mathcal{X}_T \\
			& \tilde{z}_{i+1} = \tilde{A}(s_i) \tilde{z}_i + \tilde{B}(s_i) \bar{v}_i, \\
			& (\tilde{z}_i, \bar{v}_i) \in \Tilde{\mathcal{F}}(s_i), \; i \in \mathbb{N}_0^{\bar{N}-1}.
		\end{aligned}
	\end{equation}
	where $\tilde{H}(s_i) = \sum_{j=0}^{s_i-1} (\mathbf{A}^j)^T H \mathbf{A}^j$ and 
	\begin{equation}\label{eq:F_tilde_k_explicit}
		\tilde{\mathcal{F}}(s_i) =  \bigcap_{j=0}^{s_i-1} \mathbf{A}^{-j}\bar{\mathcal{F}},
	\end{equation}
	which describes the state-input pairs that remain in $\bar{\mathcal{F}}$ for $s_i-1$ steps. Note that at time step $s_i$, the state is constrained by $\tilde{\mathcal{F}}(s_{i+1})$. An example of \eqref{eq:F_tilde_k_explicit} is shown on the left side of Fig. \ref{fig:intersectionExample}. Note that trajectories starting in $\tilde{\mathcal{F}}$ do not leave $\bar{\mathcal{F}}$ within $s_i -1$ steps.
	For polytopic sets, \eqref{eq:F_tilde_k_explicit} is computed by stacking the matrices defining the halfspaces of $\mathbf{A}^{-j}\bar{\mathcal{F}}$, in which case we arrive at the same constraint matrices as in \eqref{eq:MB_SHMPC}. However, this representation allows us to remove redundant halfspaces and thus lower the complexity of $\tilde{\mathbb{P}}$. Still, for large interval lengths $s_i$, even a non-redundant halfspace representation of $\tilde{\mathcal{F}}(s_i)$ is often still complex. In the next section, we introduce an approach to drastically reduce the constraints with only marginal loss of constraint set volume.
	\subsection{Constraint Set Approximation}
	\begin{figure}
		\centering
		\includegraphics[width=0.23\textwidth]{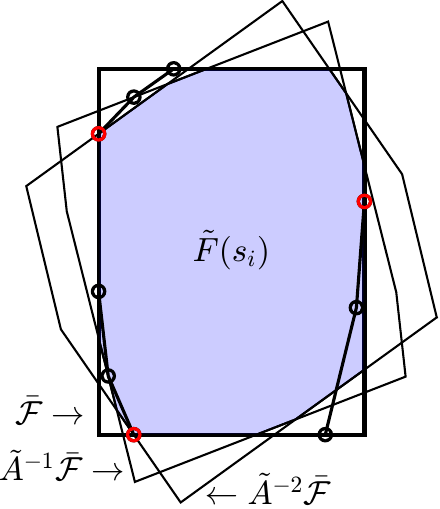}
		\includegraphics[width=0.23\textwidth]{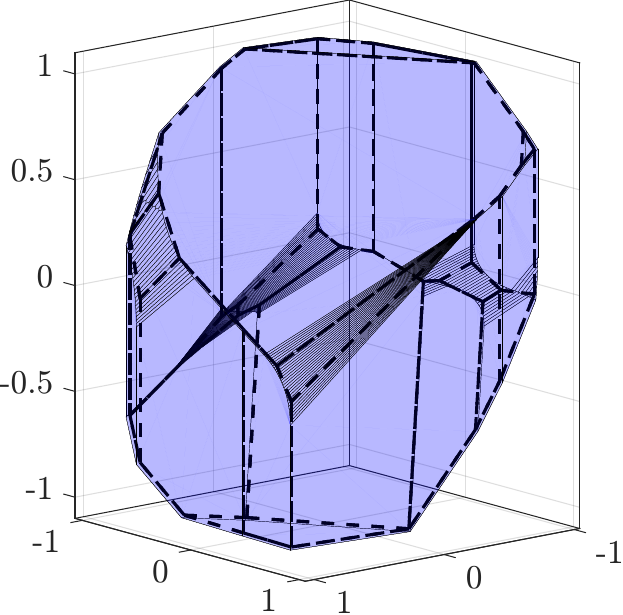}
		\caption{Left: An example of \eqref{eq:F_tilde_k_explicit} with $s_i = 3$. Red circles denote initial states. Right: $\tilde{\mathcal{F}}$ (blue) and $\tilde{\mathcal{F}}_t$ (dashed) of a randomly generated set from Example \ref{example:innerApprox_computation} for $s_i = 30$.}
		\label{fig:intersectionExample}
	\end{figure}
\begin{figure}
		\centering
		\includegraphics[width=0.48\textwidth]{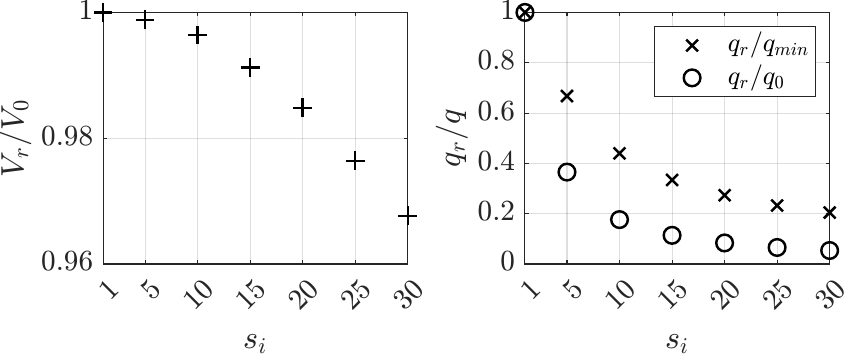}
		\caption{Ratios of volumes $V$ and number of halfspaces $q$ between $\tilde{\mathcal{F}}$ and $\tilde{\mathcal{F}}_t^*$. Subscripts $r$, $min$ and $0$ correspond to the approximation, the minimal and the non-reduced representation of $\tilde{\mathcal{F}}$, respectively.}
		\label{fig:Boxplot_F_tildeApprox}
	\end{figure}

	To further reduce $\tilde{\mathcal{F}}(s_i) = \{x \,|\, F x \leq f\}$, we construct a polytopic approximation $\tilde{\mathcal{F}}_t(s_i) = \{x \,|\, F_t x \leq f_t\}$ with fewer halfspaces. In order to ensure that all state-input pairs within $\tilde{\mathcal{F}}_t(s_i)$ also do not leave the original constraint set $\bar{\mathcal{F}}(s_i)$ within $s_i -1$ steps, the approximation needs to be a subset of $\tilde{\mathcal{F}}(s_i)$. In other words, $\tilde{\mathcal{F}}_t(s_i)$ must be an inner-approximation of $\tilde{\mathcal{F}}(s_i)$, which of course may shrink the ROA. We therefore try to maximize the volume of $\tilde{\mathcal{F}}_t(s_i)$ while enforcing $\tilde{\mathcal{F}}_t(s_i) \subseteq \tilde{\mathcal{F}}(s_i)$ via optimization. To encode set containment within a linear program (LP), we use a generalized version of Farkas lemma:
	\begin{lem}[Theorem 1 of \cite{sadraddiniLinearEncodingsPolytope2019}]\label{lem:AHPcontainment}
		Let $\mathbb{X} = \bar{x} + X \mathcal{H}_x$, $\mathbb{Y} = \bar{y} + Y \mathcal{H}_y$, where $\mathcal{H}_x = \{x \in \mathbb{R}^{n_x} | H_x x \leq h_x \}$ and $\mathcal{H}_y = \{y \in \mathbb{R}^{n_y} | H_y y \leq h_y \}$, where $q_x,q_y$ are the number of rows of $H_x$ and $H_y$, respectively. Then $\mathbb{X} \subseteq \mathbb{Y}$ if $\exists\Gamma \in \mathbb{R}^{n_y \times n_x}$, $\exists \beta \in \mathbb{R}^{n_y}$ and $\exists \Lambda \in \mathbb{R}_+^{q_y \times q_x}$ such that
		\begin{equation*}
			X = Y \Gamma, \; \bar{y} - \bar{x} = Y\beta, 
			\Lambda H_x = H_y \Gamma, \; \Lambda h_x \leq h_y + H_y \beta.
		\end{equation*}			
	\end{lem} 	
	\begin{IEEEproof}
		See Theorem 1 of \cite{sadraddiniLinearEncodingsPolytope2019} for a proof.
	\end{IEEEproof}
	The following LP uses a special case of Lemma \ref{lem:AHPcontainment} with $Y=I$ and $\bar{y}=0$ to find an optimal scaling vector $\sigma \in \mathbb{R}_+^{n+m}$ with $X = \mathrm{diag}(\sigma)$ and translation $\bar{x} \in \mathbb{R}^{n+m}$ for $\tilde{\mathcal{F}}_t(s_i)$:
	\begin{equation}\label{eq:H_in_H_opt}
		\begin{aligned}
			\min_{\bar{x}, \Lambda, \sigma} \quad & ||\sigma||_1 \\
			\textrm{s.t.} \quad & \Lambda F_t = F \mathrm{diag}(\sigma), \quad \Lambda f_t \leq f - F \bar{x}, \\
			& \Lambda \geq 0, \quad \sigma > 0.
		\end{aligned}
	\end{equation}
	Since $X$ is invertible by design, we can recover our inner-approximation as \cite{sadraddiniLinearEncodingsPolytope2019}
	\begin{equation}
		\tilde{\mathcal{F}}_t^*(s_i) = \{x \, | \, F_t \mathrm{diag}(\sigma^*)^{-1}x \leq f_t + F_t \mathrm{diag}(\sigma^*)^{-1} \bar{x}^*\},
	\end{equation}
	where the star denotes optimal values. The template $\tilde{\mathcal{F}}_t(s_i)$ can be any non-degenerate polytope of appropriate dimension. However, the choice of $\tilde{\mathcal{F}}_t(s_i)$ substantially influences the tightness and constraint reduction of the resulting approximation. In the following, we propose a useful heuristic to construct a $\tilde{\mathcal{F}}_t(s_i)$ that approximates the shape of $\tilde{\mathcal{F}}(s_i)$ efficiently. In particular, we use the structure of \eqref{eq:F_tilde_k_explicit}, but only consider a subset of the steps from $0$ to $s_i -1 $:
	\begin{equation}\label{eq:F_tilde_approx}
		\tilde{\mathcal{F}}_t(s_i) := \bigcap_{j \in \pi} \mathbf{A}^{-j}\bar{\mathcal{F}}, \; \pi \subseteq \mathbb{N}_0^{s_i-1}.
	\end{equation}
	\begin{rem}\label{rem:approx}
		With the selection of $\pi$, a trade-off between volume and constraint reduction can be achieved. The more steps there are in $\pi$, the more halfspaces $\tilde{\mathcal{F}}_t(s_i)$ contains and the closer it gets to the shape of $\tilde{\mathcal{F}}(s_i)$. We found that $\pi = \{0,\,\lfloor \frac{s_i-1}{2} \rfloor,\, s_i-1\}$, where $\lfloor \cdot \rfloor$ rounds a decimal to the nearest integer towards zero, often yields good results, even for large $s_i$.
	\end{rem}
	\begin{exmp}\label{example:innerApprox_computation}
		To assess the quality of the approximation in Remark \ref{rem:approx}, we generate random 100 second-order systems of the form \eqref{eq:system} with $\mathcal{W}=\{0\}$, discretized with a sampling time of $0.05$ seconds and constraint sets $\mathcal{F} = \mathcal{X} \times \mathcal{U}$. The state constraint set $\mathcal{X}$ is the convex hull of 62 random points while the input set $\mathcal{U} = \{u\, |\, ||u||_\infty < 1\}$ remains unchanged. All randomly generated coefficients are sampled from a uniform distribution and lie in the interval $[-1,1]$. The results are shown in Fig. \ref{fig:Boxplot_F_tildeApprox}. The volume ratio between $\tilde{\mathcal{F}}(s_i)$ and $\tilde{\mathcal{F}}_t(s_i)$ shrinks with larger $s_i$, which is expected since the number of intersection operations in \eqref{eq:F_tilde_approx} is the same, regardless of $s_i$. As a result, the number of halfspaces of $\tilde{\mathcal{F}}(s_i)$ increases with $s_i$, while the complexity of $\tilde{\mathcal{F}}_t(s_i)$ stays relatively constant. Consequently, as shown in the right plot of Fig. \ref{fig:Boxplot_F_tildeApprox}, the relative constraint reduction grows as $s_i$ increases. Note that even for large $s_i$, the constraint reduction retains over $96\%$ of the volume of $\tilde{\mathcal{F}}(s_i)$ on average. Meanwhile, constraints are reduced by $80\%$ with respect to the minimal representation and by over $94\%$ with respect to the non-reduced representation.
	\end{exmp}
	A particular set from this study for $s_i = 30$ is shown on the right side of Fig. \ref{fig:intersectionExample}. It can be seen that a lot of halfspaces stem from the coupled state-input constraints on the edges of $\tilde{\mathcal{F}}(s_i)$. Our constraint reduction approach is able to approximate this shape efficiently. 
	
	\section{Numerical Example}\label{sec:example}
	In this section, we showcase our approach with a 2-dimensional unmanned helicopter landing on an inclined surface. Though helicopter dynamics are highly nonlinear, it can be shown that with some approximations, position becomes a differentially flat output \cite{kooDifferentialFlatnessBased1999}. We assume that there exists a sufficiently fast attitude controller so that we may consider the jerk (i.e. the 3rd derivative of position) as our input. The planning model can then be described as two independent integrator chains with $x = (p_x,v_x,a_x,p_z,v_z,a_z)$ where $p$, $v$ and $a$ denote position, velocity and acceleration, respectively. Discretizing the dynamics with a sampling time of $\tau = 0.02$ seconds yields $A = \mathrm{diag}(A_x,A_z)$, $B = \mathrm{diag}(B_x,B_z)$ and $w = (w_x,w_z)$, where
	\begin{equation*}
		A_i = \begin{bmatrix}
			1 & \tau & \tau^2/2  \\ 0 & 1 & \tau \\ 0 & 0 & 1
		\end{bmatrix} B_i = \begin{bmatrix}
			\tau^3/6 \\ \tau^2 /2 \\ \tau
		\end{bmatrix}, \; w_i =  \begin{bmatrix}
			\tau^2/2 \\  \tau \\ 0
		\end{bmatrix} \bar{w}_i
	\end{equation*}
	for $i = \{x,z\}$. Attitude tracking errors, model uncertainties as well as other disturbances appear as unknown accelerations $||\bar{w}_i||_\infty \leq 0.2$. The constraints $\mathcal{F}$ are given by 
	\begin{equation*}
		\begin{gathered}
			p_z \geq 0, (-4,-10) \leq (v_x,v_z) \leq (15,5), \, -0.3v_x-v_z \leq 2 \\ ||[a_x,a_z,u_x,u_z]||_\infty \leq (4,5,3,10), \, \tan(a)p_x - p_z \leq -1
		\end{gathered}	
	\end{equation*}
	where $a = 25^{\circ}$ denotes the incline angle of the platform. The target set is given by 
	\begin{multline*}
		\mathcal{X}_T = \mathbf{R}(a)\{x \,|\,  b_l + \mathbf{g} \leq x \leq b_u + \mathbf{g}\},\\
		b_l = -(0.8,-1,1,0.9,0.4,4), \, b_u = (0.8,2.2,1,0,0.4,4)	
	\end{multline*}
	where $\mathbf{R}(a) = \mathcal{R}(a) \otimes I_3$, $\mathcal{R}(a)$ denotes the two-dimensional rotation matrix for an angle $a$ and $\mathbf{g} = (\mathbf{0}_{5\times1},9.81) (I_6 - \mathbf{R}^T(a))$. We design the ancillary controller $K$ via LQR with cost matrices $Q = \mathrm{diag}(5I_4,10I_2)$, $R = (0.1,1)$. The RPI set $\mathcal{Z}$ is computed with the method of \cite{rakovicInvariantApproximationsMinimal2005} with $\alpha = 10^{-6}$. Similar to \cite{raghuramanSetOperationsOrder2022}, we compute the RPI set using zonotopes, which provide some computational advantages but do not alter the theoretical properties of the RPI set. We simulate three different versions of $\tilde{\mathbb{P}}$ with varying constraint sets $\tilde{\mathcal{F}}(s_i)$. MPC-0 denotes $\tilde{\mathbb{P}}$ with the full representation of $\tilde{\mathcal{F}}(s_i)$, which is equivalent to solving $\mathbb{P}$. Similarly, MPC-min and MPC-a represent $\tilde{\mathbb{P}}$ with the minimal representation of the constraint set and with its low-complexity approximation $\tilde{\mathcal{F}}_t(s_i)$, respectively. To penalize deviation from a setpoint $x_r = (-0.7,1.4,0.2,-0.4,-4.1,-0.9)$, we consider 
	\begin{equation*}
		\sum_{i=0}^{\bar{N}-1} (z_i-x_r,v_i)^T\tilde{H}(s_i)(z_i-x_r,v_i) + (z_{\bar{N}}-x_r)^TP(z_{\bar{N}}-x_r)
	\end{equation*}
	as our cost function with $H = \mathrm{diag}(Q,R)$ and $P$ as the corresponding solution to the discrete-time Riccati equation. The initial horizon length is set to $N_0 = 300$ with a maximum number of decision inputs $\bar{N}_{\mathrm{max}} = 10$ and initial blocking matrix $M_0 = \mathbf{1}_{30\times1} \otimes I_{\bar{N}}$. When applying $\Psi(M_k)$ using Proposition \ref{prop:feasibleSplits}, we choose $j^* = \arg \max_j s_j$ and $i = s_{j^*}/2$, i.e. we split the largest blocking interval through the middle. When $s_{j^*}$ is odd, we round $i$ to the nearest integer towards infinity. If there exist multiple possible $j^*$, we choose the largest one.
	
\begin{figure}
		\centering
		\includegraphics[width=0.48\textwidth]{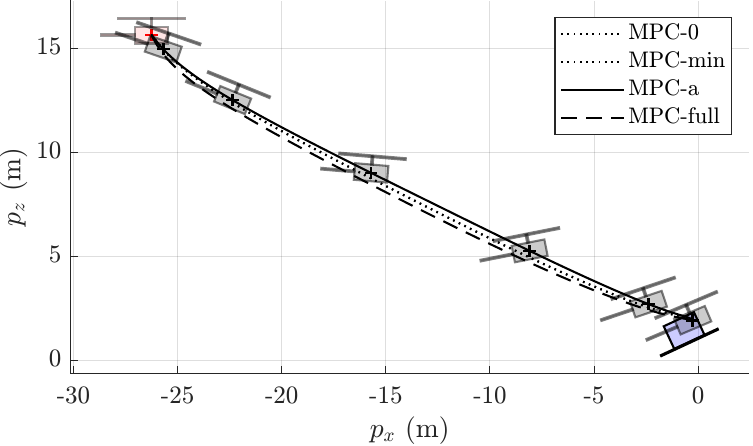}
		\caption{Closed-loop simulation results of a helicopter landing. For MPC-a, the attitude and center of mass are plotted every second with the initial condition marked in red. The blue region denotes the projection of the target set. MPC-0 and MPC-min lie on top of each other.}
		\label{fig:helilanding}
	\end{figure}

	Since the number of blocked inputs is highest at the initial condition, we first compute the trajectory costs $J$ and computation times $t_c$ for nominal open-loop trajectories planned at $k=0$. We benchmark MPC-0, MPC-min and MPC-a with a non-blocked SHMPC denoted by MPC-full. The averaged results of 50 randomly generated feasible initial conditions are shown in the open-loop column of Tab. \ref{tab:results}. Afterwards, the MPC algorithms are simulated in closed-loop. Their trajectories are shown in Fig. \ref{fig:helilanding} and their costs and computation times are reported in the closed-loop column of Tab. \ref{tab:results}.
	Overall, the costs of blocked trajectories are only slightly larger than non-blocked costs. As expected, costs for MPC-min and MPC-0 are equal since in MPC-min, only redundant halfspaces are removed. Though MPC-a uses approximative constraint sets, its cost differs marginally from MPC-min and MPC-0 while reducing computation times by orders of magnitude. Note also that the cost ratio in closed-loop is smaller than in open-loop. This showcases the efficacy of utilizing free inputs to split blocking intervals and recovering some optimality when getting closer to the target. All computations are run in MATLAB on a laptop with an Intel Core i7-11850H using the solver osqp \cite{stellatoOSQPOperatorSplitting2020}.
	\begin{table}
		\centering	
		\caption{Trajectory costs and computation times}
		\begin{tabular}{l|cc|cc}
			& \multicolumn{2}{c}{open-loop} & \multicolumn{2}{|c}{closed-loop} \\
			& $J/J_{full}$ & $t_c$ (s) & $J/J_{full}$ & $t_c$ (s) \\
			\hline
			MPC-0 & 1.027 & 0.105 & 1.002 & 0.102 \\
			MPC-min & 1.027 & 0.041 & 1.002 & 0.037 \\
			MPC-a & 1.027 & 0.004 & 1.004 & 0.005  \\
			\hline
			MPC-full & 1 & 1.541 & 1 & 2.389
		\end{tabular}
		\label{tab:results}
	\end{table}
	\section{Conclusion and Outlook}\label{sec:conclusion}
	In this paper, we proposed a shrinking horizon Model Predictive Control approach with move blocking that is scalable to large prediction horizons. A function is designed that generates blocking matrices which ensure recursive feasibility while efficiently utilizing the available number of decision inputs. Furthermore, the optimal control problem is reformulated in a way that enables the reduction of the constraint set complexity. A low-complexity inner approximation is then obtained thorugh optimization. The computational efficiency of the approach is demonstrated with a helicopter landing. In the future, we plan to confirm the computational efficiency in hardware experiments and extend the approach to continuous systems.
	
	\bibliographystyle{IEEEtran}
	\bibliography{bibliography.bib}

\begin{thebibliography}{10}
\providecommand{\url}[1]{#1}
\csname url@samestyle\endcsname
\providecommand{\newblock}{\relax}
\providecommand{\bibinfo}[2]{#2}
\providecommand{\BIBentrySTDinterwordspacing}{\spaceskip=0pt\relax}
\providecommand{\BIBentryALTinterwordstretchfactor}{4}
\providecommand{\BIBentryALTinterwordspacing}{\spaceskip=\fontdimen2\font plus
\BIBentryALTinterwordstretchfactor\fontdimen3\font minus
  \fontdimen4\font\relax}
\providecommand{\BIBforeignlanguage}[2]{{%
\expandafter\ifx\csname l@#1\endcsname\relax
\typeout{** WARNING: IEEEtran.bst: No hyphenation pattern has been}%
\typeout{** loaded for the language `#1'. Using the pattern for}%
\typeout{** the default language instead.}%
\else
\language=\csname l@#1\endcsname
\fi
#2}}
\providecommand{\BIBdecl}{\relax}
\BIBdecl

\bibitem{mayneModelPredictiveControl2014}
D.~Q. Mayne, ``Model predictive control: {{Recent}} developments and future
  promise,'' \emph{Automatica}, vol.~50, no.~12, pp. 2967--2986, 2014.

\bibitem{erenModelPredictiveControl2017}
U.~Eren, A.~Prach, B.~B. Koçer, S.~V. Raković, E.~Kayacan, and
  B.~Açıkmeşe, ``Model {{Predictive Control}} in {{Aerospace Systems}}:
  {{Current State}} and {{Opportunities}},'' \emph{Journal of Guidance,
  Control, and Dynamics}, vol.~40, no.~7, pp. 1541--1566, 2017.

\bibitem{mayneConstrainedModelPredictive2000}
D.~Mayne, J.~Rawlings, C.~Rao, and P.~Scokaert, ``Constrained model predictive
  control: {{Stability}} and optimality,'' \emph{Automatica}, vol.~36, no.~6,
  pp. 789--814, 2000.

\bibitem{richardsRobustVariableHorizon2006}
A.~Richards and J.~P. How, ``Robust variable horizon model predictive control
  for vehicle maneuvering,'' \emph{Int. J. Robust Nonlinear Control}, vol.~16,
  no.~7, pp. 333--351, 2006.

\bibitem{richardsModelPredictiveControl2003}
A.~Richards and J.~How, ``Model predictive control of vehicle maneuvers with
  guaranteed completion time and robust feasibility,'' in \emph{2003 American
  Control Conference (ACC)}, vol.~5, 2003, pp. 4034--4040.

\bibitem{shekharRobustVariableHorizon2012}
R.~C. Shekhar and J.~M. Maciejowski, ``Robust variable horizon {{MPC}} with
  move blocking,'' \emph{Systems \& Control Letters}, vol.~61, no.~4, pp.
  587--594, 2012.

\bibitem{shekharRobustModelPredictive2015}
R.~C. Shekhar, M.~Kearney, and I.~Shames, ``Robust {{Model Predictive Control}}
  of {{Unmanned Aerial Vehicles Using Waysets}},'' \emph{Journal of Guidance,
  Control, and Dynamics}, vol.~38, no.~10, pp. 1898--1907, 2015.

\bibitem{koelnTwoLevelHierarchicalMissionBased2018}
J.~P. Koeln and A.~G. Alleyne, ``Two-level hierarchical mission-based model
  predictive control,'' in \emph{2018 American Control Conference (ACC)}, 2018,
  pp. 2332--2337.

\bibitem{farahaniShrinkingHorizonModel2017}
S.~S. Farahani, R.~Majumdar, V.~S. Prabhu, and S.~E.~Z. Soudjani, ``Shrinking
  horizon model predictive control with chance-constrained signal temporal
  logic specifications,'' in \emph{2017 American Control Conference (ACC)},
  2017, pp. 1740--1746.

\bibitem{cagienardMoveBlockingStrategies2007}
R.~Cagienard, P.~Grieder, E.~Kerrigan, and M.~Morari, ``Move blocking
  strategies in receding horizon control,'' \emph{Journal of Process Control},
  vol.~17, no.~6, pp. 563--570, 2007.

\bibitem{mayneRobustModelPredictive2005}
D.~Mayne, M.~Seron, and S.~Raković, ``Robust model predictive control of
  constrained linear systems with bounded disturbances,'' \emph{Automatica},
  vol.~41, no.~2, pp. 219--224, 2005.

\bibitem{chisciSystemsPersistentDisturbances2001}
L.~Chisci, J.~Rossiter, and G.~Zappa, ``Systems with persistent disturbances:
  Predictive control with restricted constraints,'' \emph{Automatica}, vol.~37,
  no.~7, pp. 1019--1028, 2001.

\bibitem{sadraddiniLinearEncodingsPolytope2019}
S.~Sadraddini and R.~Tedrake, ``Linear encodings for polytope containment
  problems,'' in \emph{IEEE 58th Conference on Decision and Control (CDC)},
  2019, pp. 4367--4372.

\bibitem{kooDifferentialFlatnessBased1999}
T.~Koo and S.~Sastry, ``Differential flatness based full authority helicopter
  control design,'' in \emph{IEEE 38th Conference on Decision and Control
  (CDC)}, vol.~2, 1999, pp. 1982--1987.

\bibitem{rakovicInvariantApproximationsMinimal2005}
S.~Rakovic, E.~Kerrigan, K.~Kouramas, and D.~Mayne, ``Invariant approximations
  of the minimal robust positively {{Invariant}} set,'' \emph{IEEE Transactions
  on Automatic Control}, vol.~50, no.~3, pp. 406--410, 2005.

\bibitem{raghuramanSetOperationsOrder2022}
V.~Raghuraman and J.~P. Koeln, ``Set operations and order reductions for
  constrained zonotopes,'' \emph{Automatica}, vol. 139, 2022, 110204.

\bibitem{stellatoOSQPOperatorSplitting2020}
B.~Stellato, G.~Banjac, P.~Goulart, A.~Bemporad, and S.~Boyd, ``{{OSQP}}: An
  operator splitting solver for quadratic programs,'' \emph{Mathematical
  Programming Computation}, vol.~12, no.~4, pp. 637--672, 2020.

\end{thebibliography}
	
\end{document}